\documentclass[sigconf,authoryear]{acmart}

\setcopyright{rightsretained}
\copyrightyear{2026}
\acmConference[EvalMG '26]{The 2nd Workshop on Evaluation for Multimodal Generation @ SIGIR 2026}{July 24, 2026}{Melbourne, Australia}
\acmYear{2026}
\acmISBN{}
\acmDOI{}
\usepackage{microtype}
\usepackage{graphicx}
\usepackage{tikz}
\usetikzlibrary{arrows.meta,positioning,fit,backgrounds,calc}
\usepackage{booktabs}
\usepackage{multirow}
\usepackage{makecell}
\usepackage{xcolor}

\usepackage{amsmath,amssymb}
\usepackage{enumitem}
\usepackage[capitalize,noabbrev]{cleveref}

\begin{document}

\title{Chains That See, Answers That Don't: A Multi-Aspect Evaluation Recipe for Forced Chain-of-Thought on Video-MME}

\author{Zhichao Fan}
\affiliation{%
  \institution{University of Illinois Urbana-Champaign}
  \city{Urbana}
  \state{Illinois}
  \country{USA}
}
\email{zhichao8@illinois.edu}

\author{Yanhang Li}
\affiliation{%
  \institution{Northeastern University}
  \city{Boston}
  \state{Massachusetts}
  \country{USA}
}
\email{li.yanha@northeastern.edu}

\author{Zexin Zhuang}
\affiliation{%
  \institution{Southern Methodist University}
  \city{Dallas}
  \state{Texas}
  \country{USA}
}
\email{zexinz@smu.edu}

\renewcommand{\shortauthors}{Fan, Li, and Zhuang}
\renewcommand{\shorttitle}{Forced CoT on Video-MME: a three-probe recipe}

\begin{abstract}
Forced chain-of-thought (CoT) is widely assumed to make vision--language models more reliable on video question answering. We propose a small three-probe evaluation recipe to test that assumption: paired accuracy across direct, CoT, answer-first, and no-video conditions; a counterfactual video-swap diagnostic over the CoT chains; and a four-rung visual-degradation ladder. Each probe is reported under both a strict and a permissive regex scorer, with multiplicity correction over a manuscript-declared primary family.

Applied to Qwen2.5-VL on Video-MME subsets, the recipe returns a two-part finding. The CoT chains \emph{are} strongly video-conditioned: swapping the input video collapses chain overlap and flips most final letters, the opposite of what a ``boilerplate-chain'' null would predict. Yet on the same data, forced CoT does not improve MCQ accuracy, and on the smaller (7B) model it produces a small but statistically supported drop under a post-hoc primary scorer choice (the same effect is directionally similar but not Holm-surviving under the alternative scorer). We do not claim this generalizes beyond the Qwen2.5-VL / Video-MME instantiation; the raw responses and a single recomputation script will be released with the supplementary material so every number can be re-derived.
\end{abstract}

\ccsdesc[500]{Computing methodologies~Natural language processing}
\ccsdesc[500]{Computing methodologies~Computer vision tasks}
\ccsdesc[300]{Computing methodologies~Model verification and validation}

\keywords{evaluation methodology; chain-of-thought; video question answering; multimodal generation; vision-language models; benchmark probes; multiple-comparisons correction}

\maketitle

\section{Introduction}
\label{sec:intro}

Forced chain-of-thought (CoT) prompting---instructing the model to produce several reasoning steps before committing to an answer---has become a default setting when adapting vision--language models (VLMs) to multiple-choice benchmarks. A natural assumption is that these written chains are functional: they consult the visual input and steer the final answer. The competing prediction---what we call the ``boilerplate-chain'' null---is that forced CoT chains can be largely input-invariant (chain token-Jaccard near $1.0$ across input swaps; final letter rarely flips) yet still shift the final answer through format effects, regardless of whether the model meaningfully ``reads'' the input. We are aware of unpublished concurrent observations consistent with this null in some small-VLM image-VQA settings, but do not rely on them quantitatively; we instead test the prediction directly on Video-MME with Qwen2.5-VL, which goes the opposite direction. Video-MME is a useful measurement stress-test for that prediction because it stratifies $2700$ MCQ items across $12$ task types and three duration buckets, includes a parse-rate-controllable trailing ``Answer:~X'' format, and has both no-video and same-task-different-video counterfactuals available without any new data collection.

Two evaluation questions follow. (i)~Does a clean ``CoT helps / does not help'' verdict require more than paired accuracy with a format control? (ii)~Are reasoning-chain claims (input-conditioned, input-invariant) testable independently of accuracy claims? We argue both require a small but specific set of probes.

\paragraph{Contributions.}
\begin{enumerate}\setlength\itemsep{0em}
\item \textbf{A three-probe evaluation recipe for video-CoT.} The recipe comprises (a)~paired primary accuracy with a $\{$direct, CoT, AF, no-video$\}$ design under \emph{both} permissive $(a$--$d)$ and strict $(a$--$c)$ scoring, with exact McNemar and 50k paired-bootstrap CIs; (b)~a counterfactual video-swap probe scoring chain token-Jaccard, paired-parsed flip rate, and original-label retention with parsed-same / parsed-different / strict-no-parse decomposition; (c)~a four-rung visual-degradation branch (real $\to$ shuffle $\to$ single $\to$ black) with an exact-permutation Spearman test, and two contamination conditions (swap-task, swap-domain) reported on a separate axis. Each probe targets a distinct evaluation failure mode. We do not claim a settled \emph{protocol}: the recipe is instantiated once on one benchmark and one model family.
\item \textbf{Evidence against a boilerplate-chain signature on Video-MME with Qwen2.5-VL.} Forced CoT chains here \emph{do} depend on the visual input in this instantiation. Under a same-task video swap, chain token-Jaccard is $0.140$/$0.119$ (32B/7B) and the final letter flips on $67.2\%$/$59.9\%$ of paired-parsed items (permissive; $63.3\%$/$59.4\%$ strict, smaller pair). Decoding is greedy, so this divergence is not stochasticity.
\item \textbf{Video-conditioned chains do not imply MCQ utility in this instantiation, and on 7B they hurt.} Under our (post-hoc) primary strict scorer, raw CoT$-$direct is $-5.33$ pp on 32B ($p{=}0.048$, Holm $0.125$) and $\mathbf{-7.32}$ \textbf{pp on 7B ($p{=}0.003$, Holm-adjusted $p{=}0.012$, the only Holm-surviving effect under our post-hoc primary scoring choice over the four-test primary family)}. Permissive raw is $-2.33$/$-5.81$ pp ($p{=}0.43$/$0.019$, Holm $0.43$/$0.077$); the 7B harm is directionally similar but does not Holm-survive under permissive scoring, so the strict-as-primary choice is what makes the headline ``Holm-surviving'' phrasing apply (\cref{sec:setup}). The 32B parsed-only sign flips between scorers (permissive $-2.01$, strict $+3.05$ on a smaller all-four-parsed subset)---we read this as a parse-rate selection effect rather than a 32B improvement. An answer-first control rules out a pure answer-position effect under permissive scoring but its scope narrows under strict.
\end{enumerate}

The reliability-themed message for evaluators is that \emph{grounded reasoning traces can fail to improve task accuracy}. Sensitivity of chains to the visual input is not a sufficient condition for downstream gains, and the two should be measured separately.

\section{Related Work}
\label{sec:related}

\textbf{CoT prompting.} \citet{weicot2022} introduced explicit chain-of-thought prompting as a way to elicit multi-step reasoning from large language models; \citet{kojimazsot2022} showed a zero-shot ``Let's think step by step'' variant succeeds on many reasoning tasks. Both are text-modal. Applying forced CoT to vision--language benchmarks is now routine. Related vision and video work has used chain-of-thought language prompting for segmentation, instruction tuning for motion-language models, and LLMs for repetitive-action counting~\citep{li2024image,li2025human,yao2025countllm}.

\textbf{CoT faithfulness and skepticism.} A growing line of work questions whether stated reasoning chains reflect the computation the model actually performs. \citet{turpin2023} demonstrate that LLMs rationalize answers already driven by spurious features like option ordering. \citet{lanham2023} propose paraphrase and corruption protocols that reveal frequent chain--answer inconsistency. \citet{madaan2022text} show superficial lexical patterns of chains can drive downstream behavior even when semantic content is degraded. Adjacent work targets narrower pieces of this measurement problem: \citet{cai2025role} analyze deductive and inductive reasoning modes in LLMs, while explanation-consistency and reasoning-trace audit studies reinforce the need to measure generated reasoning artifacts rather than treat them as transparent evidence~\citep{lan-etal-2025-attention,li2026auditing}. Our work is in the same skeptical tradition but asks a complementary question: given that a chain \emph{is} strongly input-conditioned, does it therefore improve the final answer? The data say no.

\textbf{CoT on images and videos.} The boilerplate-chain prediction---that forced CoT chains can be largely input-invariant under input perturbation while still shifting the final letter through format effects---has been informally observed in some small-VLM image-VQA settings (we are aware of unpublished concurrent observations of this kind but do not rely on them quantitatively). The present paper tests the prediction on (a)~video and (b)~a 32B-scale model. We find the prediction refuted on this setup, while the no-gain pattern partially survives at 32B and 7B exhibits Holm-surviving harm under our (post-hoc) primary scorer. \citet{zhang2023multimodalcot} and follow-on work apply CoT to multimodal benchmarks; our setting differs in that we treat the chain itself as an object of evaluation rather than as a means to a higher score.

\textbf{Evaluation methodology for multimodal generation.} Our recipe contributes to a thread that treats benchmark scores as artifacts of measurement design rather than direct reflections of capability. Adjacent diagnostic benchmarks for visual RAG and text-to-image generation make a similar methodological move by evaluating specific failure modes and bias dimensions rather than only reporting aggregate task scores~\citep{ji2025mrag,luo2024bigbench,luo2026biasig,luo2024faintbench}. Recent benchmark-audit and reproducibility studies further emphasize predeclared detectable effects and configuration sensitivity~\citep{zhuang2026preregistering,li2026safetyrepro}; RAG diagnostics provide parallel evidence-facing checks by testing context compliance under knowledge conflict~\citep{chen2026doesragknowretrieval} and whether cited evidence warrants the answer~\citep{qian2026relevantwarrantedevidenceforcecalibration}. Equivalence testing~\citep{schuirmann1987tost,lakens2017tost} and parse-rate-controlled scoring are standard in clinical and psycholinguistic measurement but remain rare in VLM evaluation, where ``CoT $-$ direct $<0$ but $p{>}0.05$'' is routinely reported as ``no effect'' rather than as an underpowered null. Counterfactual input substitution and graded-degradation ladders are likewise underused as input-conditioning checks for generation-style outputs.

\textbf{Video MCQ.} Video-MME~\citep{videomme2024} is a test-only benchmark of $2700$ multiple-choice questions over $900$ videos, stratified across $12$ task types and $3$ duration buckets. Its MCQ format lets us evaluate CoT accuracy directly without rubric-grading.

\textbf{Models.} We use the Qwen2.5-VL family~\citep{qwen25vl2025}, which supports arbitrary-frame video input through a unified visual encoder. We evaluate the $7$B-Instruct variant in full precision (fp16) and the $32$B-Instruct variant~\citep{qwen25vl32b2025} under \texttt{bitsandbytes} NF4 4-bit quantization; both checkpoints are public.

\section{Evaluation Protocol}
\label{sec:setup}

\begin{figure*}[t]
    \centering
    \input{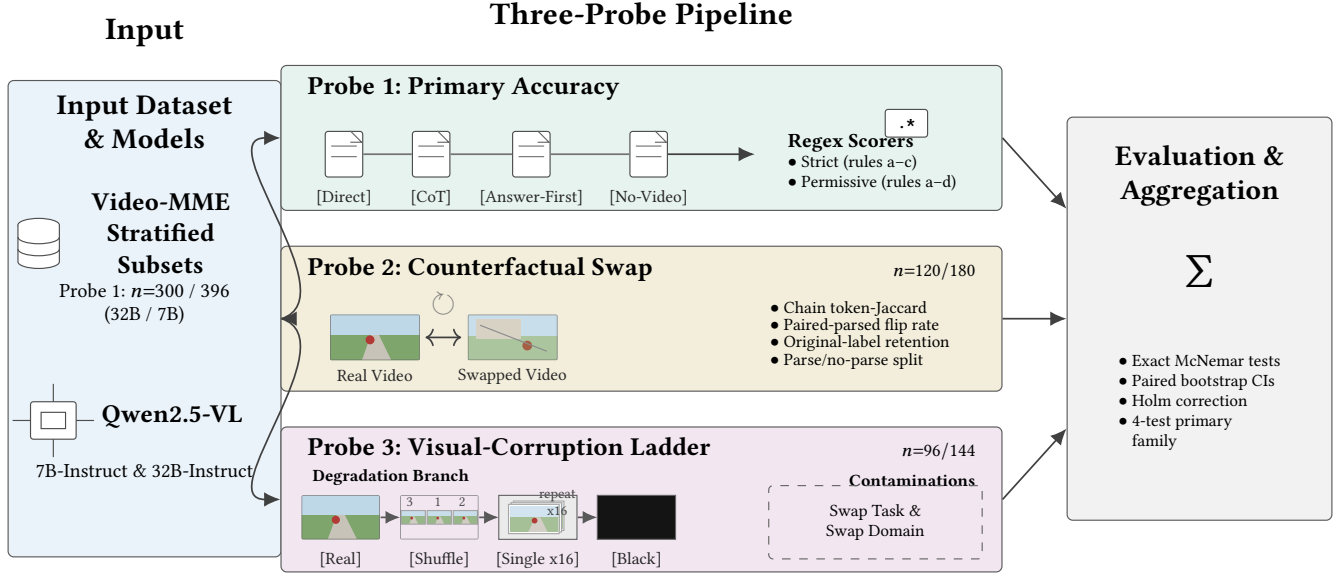}
    \caption{Three-probe evaluation pipeline. Stratified Video-MME subsets feed Probe 1 (paired direct, CoT, answer-first, and no-video accuracy under strict and permissive regex scorers; $n{=}300/396$), Probe 2 (counterfactual video-swap diagnostic over CoT chains; $n{=}120/180$), and Probe 3 (four-rung visual-degradation ladder plus two contamination conditions reported on a separate axis; $n{=}96/144$). Outputs are aggregated with exact McNemar and Holm correction over a manuscript-declared four-test primary multiplicity family.}
    \Description{Flow diagram with three probes branching from stratified Video-MME subsets: primary accuracy conditions, counterfactual swap diagnostics, and a visual degradation and contamination ladder, followed by statistical aggregation.}
    \label{fig:pipeline}
\end{figure*}

\paragraph{Benchmark and scope.} Video-MME contains $2700$ multiple-choice questions over $900$ videos partitioned into three duration buckets (short, medium, long) and twelve task types. The three-probe evaluation pipeline (Figure~\ref{fig:pipeline}) instantiates on a stratified subset: $25$ questions per task type for 32B ($n{=}300$) and $33$ per task type for 7B ($n{=}396$). A full-benchmark evaluation is deferred (see \cref{sec:limitations}).

\paragraph{Models and inference.} We run Qwen2.5-VL-7B-Instruct in \texttt{bf16}/\texttt{fp16} and Qwen2.5-VL-32B-Instruct under NF4 4-bit \texttt{bnb} quantization (visual module in full precision). Inference: HuggingFace Transformers, SDPA attention, \textbf{greedy decoding} (\texttt{do\_sample=False}), $16$ uniformly sampled frames per video, \texttt{max\_side=448}. Greedy decoding matters for the swap probe (\cref{sec:swap}): low chain overlap is not sampling noise.

\paragraph{Probe 1 --- Primary conditions.} The same $300$ / $396$ questions are evaluated under four prompt conditions (full prompt templates in \cref{app:prompts}):

\begin{center}\footnotesize
\begin{tabular}{@{}lp{5.0cm}l@{}}
\toprule
Condition & Prompt tail (abbreviated) & Video \\
\midrule
direct       & ``Answer with only the letter.'' & real frames \\
CoT          & ``$\geq 5$ numbered steps, each citing a frame, then Answer: X.'' & real frames \\
answer-first & ``Answer: X first, then 5 numbered steps.'' & real frames \\
no-video     & same as \textit{direct} & $0$ frames \\
\bottomrule
\end{tabular}
\end{center}

\paragraph{Probe 2 --- Counterfactual swap.} $120$ (32B) / $180$ (7B) stratified questions are re-evaluated under the CoT condition with the real video replaced by a swap. The swap candidate pool is filtered to same task type and different \texttt{videoID}; within that pool we prefer same duration bucket. A single swap is drawn per question with a seeded RNG (\texttt{seed=0}). The probe outputs four quantities per pair: chain token-Jaccard, paired-parsed flip rate, original-label retention under invalid video (a refusal-vs-coincidental-match diagnostic, not a valid score), and a parsed-same / parsed-different / strict-no-parse decomposition. Token-Jaccard is computed on lower-cased whitespace tokens after stripping the trailing \texttt{Answer:~X} line and a closed stopword/filler list (English stopwords plus ``video'', ``frame(s)'', ``image(s)'', ``shows'', ``sees'', ``observe(d)''); no stemming. Full token-list is in the supplementary recomputation script.

\paragraph{Probe 3 --- Visual-corruption ladder.} $96$ (32B) / $144$ (7B) questions are evaluated with CoT under six visual-input conditions: a \emph{degradation} branch \{\texttt{real} $\to$ \texttt{shuffle} (frames randomly permuted) $\to$ \texttt{single} (middle frame repeated $16$ times) $\to$ \texttt{black} (all-black frames)\} of monotonically decreasing visual signal, plus two \emph{contamination} conditions \{\texttt{swap\_task} (different video, same task type), \texttt{swap\_domain} (different video, different top-level Video-MME domain, e.g., \emph{Knowledge} vs.\ \emph{Sports}; see \cref{app:prompts})\} which are not on the same scalar severity axis. The same question text and answer options are kept fixed across the six conditions. We report Spearman $\rho_4$ across the four-rung degradation branch only (with an exact-permutation $p$-value over $4!{=}24$ permutations); we do not report a $p$-value for any six-condition concatenation because the contamination rungs are not on the same severity axis as the degradation rungs.

\paragraph{Subtitle ablation.} A supplementary ablation on 32B with $n{=}61$ paired (limited by SRT availability) is reported in \cref{app:subtitle}.

\paragraph{Scoring (definitions only).} A deterministic regex cascade extracts a single letter from each response. The \emph{strict} scorer (\textbf{primary}) applies three rules: (a)~an explicit ``Answer:~X'' / ``final answer:~X'' tag (or Chinese equivalents), X is a word-bounded \texttt{[ABCDabcd]}; (b)~a line-initial isolated word-bounded letter followed by \texttt{)} / \texttt{.} / \texttt{:}; (c)~a word-bounded isolated letter at end of response. The \emph{permissive} scorer (sensitivity contrast) appends rule (d)~the last word-bounded A/B/C/D anywhere, which can coerce refusal-style outputs (e.g.,~``the video does not match any option'') into a letter. We always report both. Unparsed responses count as wrong in raw and do not contribute to parsed-only metrics. Until item-level agreement with the official Video-MME evaluator is established, no claim in this paper is scorer-independent; we treat ``strict regex'' and ``permissive regex'' as bracketing operationalizations rather than as the official Video-MME score. \emph{All numerical results---primary effects, scorer comparisons, sign flips---are deferred to \cref{sec:results}; this section defines instruments only.}

\paragraph{Statistics.} Exact two-sided McNemar~\citep{mcnemar1947note} for paired difference, with mid-$p$ and asymptotic alternatives treated as known variants rather than our primary test~\citep{fagerland2013recommended}; paired bootstrap percentile CIs~\citep{efron1993bootstrap} on $50{,}000$ resamples; Holm sequentially-rejective correction~\citep{holm1979simple} for the four-test primary multiplicity family (CIs reproducible to within $\pm 0.05$ pp at this $n$). For the 32B null we report a $\pm 5$ pp \emph{post-hoc} sensitivity margin and the $90\%$ CI; we do not call this a TOST equivalence test because the margin was not pre-specified. For the visual-corruption ladder we report Spearman $\rho$ across the four-rung degradation branch with a \emph{one-tailed} exact-permutation $p$-value in the monotone-decreasing direction (over $4!{=}24$ permutations); we do not report a $p$-value across the six-condition concatenation because two of the six conditions (contamination) are not on the same severity axis as the four degradation rungs. Random seed is fixed ($=\!0$) for all sampling and bootstrap. \emph{One-source rule:} every number in this paper is computed by \texttt{scripts/recompute\_canonical.py} from the supplementary JSONL responses; no hand-curated aggregates are quoted.

\paragraph{Parsed-only estimands.} Two parsed-only estimands are well-defined: \emph{pair-parsed} (both endpoints of the contrast parse) and \emph{all-four-parsed} (all four conditions parse). Permissive: all-four-parsed $n{=}299$/$391$ (32B/7B). Strict: $n{=}262$/$343$, because CoT and AF have lower strict parse rates (\cref{app:parse}). We report \emph{both} so the scorer choice is fully exposed; strict is the primary scorer for sign stability since rule~$(d)$ can coerce refusal-style responses into letters.

\paragraph{Primary multiplicity family (manuscript-declared, not preregistered).} This study was not preregistered, so we cannot label any test ``confirmatory'' on a timestamped basis. We instead manuscript-declare a \emph{primary multiplicity family} of four tests for Holm correction: raw paired-McNemar direct-vs-CoT under the strict scorer for each model, plus the ladder Spearman exact-permutation test for each model. The choice of strict as primary follows from the rule~$(d)$ refusal-coercion concern (\cref{tab:strict-swap}); we report the permissive-scorer family as a sensitivity contrast in \cref{tab:primary} and explicitly disclose that \emph{the strict-as-primary choice is what makes the 7B harm Holm-survive at $\alpha{=}0.05$}; under permissive scoring the 7B harm is directionally similar but not Holm-surviving. All other reported $p$-values, CIs, equivalence margins, AF-vs-direct contrasts, contamination conditions, per-task breakdowns, video-dependent subset, and subtitle ablation are \emph{exploratory} and are reported as point estimates with CIs, not as significance tests.

\paragraph{Honesty flags.} We anticipate four reader concerns up front. (i)~The \texttt{no\_video} floor is $42\%$ (32B) / $37\%$ (7B) against a $25\%$ random baseline, indicating question-only solvability; we therefore also report CoT-vs-direct on the \emph{video-dependent subset} (items \texttt{no\_video} gets wrong, \cref{app:vds}). (ii)~Decoding is deterministic, so low chain-Jaccard under swap is not output stochasticity. (iii)~The answer-first condition rules out a \emph{pure} answer-position account but does not control for output length, instruction complexity, or compliance burden; we report it as the narrow control it is. (iv)~The custom regex extractor is not the official Video-MME evaluator; we mitigate by reporting both permissive $(a)$--$(d)$ and strict $(a)$--$(c)$ scorers on every primary table, but no claim in this paper should be read as scorer-independent until item-level agreement with the official evaluator is established.

\section{Results}
\label{sec:results}

\subsection{Does forced CoT help MCQ accuracy?}
\label{sec:primary}

\Cref{tab:primary} is our main result. All numbers are recomputed by one script (\texttt{recompute\_canonical.py}) from the supplementary JSONL responses, under both the permissive $(a)$--$(d)$ and strict $(a)$--$(c)$ regex cascades; strict drops the permissive last-letter-anywhere fallback (\cref{sec:setup}).

\begin{table}[t]
\centering\footnotesize
\caption{Primary paired direct-vs-CoT accuracy on Video-MME, both models, both scorers (perm.\ = permissive $(a)$--$(d)$; strict $(a)$--$(c)$, primary). \emph{Raw} counts unparsed as wrong; \emph{all4} restricts to items where every condition parses under the named scorer. CIs: $50{,}000$-resample paired bootstrap. McNemar exact two-sided. Holm $p$ is over each scorer's four-test family (raw McNemar $\times$ 2 models + ladder Spearman $\times$ 2 models, \cref{sec:setup}); strict is the primary family. AF$-$direct and no-video are reported in prose below.}
\label{tab:primary}
\setlength{\tabcolsep}{4pt}
\begin{tabular}{@{}llcccc@{}}
\toprule
Model & Subset (scorer; $n$) & dir / CoT (\%) & $\Delta$pp & 90\% CI & McN. / Holm \\
\midrule
\multirow{4}{*}{\makecell[l]{Qwen\\32B}}
 & raw (perm.; $n{=}300$)        & $61.67 / 59.33$ & $-2.33$ & $[-6.33,+1.67]$ & $0.43 / 0.43$ \\
 & raw (strict; $n{=}300$)       & $61.67 / 56.33$ & $-5.33$ & $[-9.33,-1.33]$ & $0.048 / 0.125$ \\
 & all4 (perm.; $n{=}299$)       & $61.54 / 59.53$ & $-2.01$ & $[-6.02,+2.01]$ & $0.50$ \\
 & all4 (strict; $n{=}262$)      & $61.45 / 64.50$ & $+3.05$ & $[-0.76,+6.87]$ & $0.23$ \\
\midrule
\multirow{4}{*}{\makecell[l]{Qwen\\7B}}
 & raw (perm.; $n{=}396$)        & $59.09 / 53.28$ & $-5.81$ & $[-9.60,-2.02]$ & $0.019 / 0.077$ \\
 & raw (strict; $n{=}396$)       & $59.09 / 51.77$ & $\mathbf{-7.32}$ & $[-11.11,-3.28]$ & $\mathbf{0.003} / \mathbf{0.012}$ \\
 & all4 (perm.; $n{=}391$)       & $59.34 / 53.45$ & $-5.88$ & $[-9.72,-2.05]$ & $0.018$ \\
 & all4 (strict; $n{=}343$)      & $59.77 / 54.23$ & $-5.54$ & $[-9.62,-1.46]$ & $0.034$ \\
\bottomrule
\end{tabular}
\end{table}

\noindent\textbf{Same-subset secondary numbers.}
\textbf{AF$-$direct (pp):} 32B $\approx +2$ on every row (all 90\% CIs include zero); 7B $-0.76$/$-3.79$/$-0.51$/$+1.16$ on raw-perm./raw-strict/all4-perm./all4-strict. Only 7B raw-strict $-3.79$ pp excludes zero, so the AF control is narrower under strict on 7B (some 7B AF outputs are caught by fallback $(d)$ but not by $(a)$--$(c)$).
\textbf{no-video accuracy:} 32B $\approx 42$, 7B $\approx 37$, both well above the $25\%$ four-way random baseline---hence the video-dependent subset analysis in \cref{app:vds}.

Three observations follow. \textbf{First, on 32B the result is scorer-sensitive in two distinct ways.} (a)~The raw-strict effect is $-5.33$ pp ($p{=}0.048$, Holm $0.125$) while raw-permissive is $-2.33$ pp ($p{=}0.43$); the difference is $3.0$ pp of CoT responses that the permissive fallback~$(d)$ coerces to a (mostly correct) letter. (b)~Conditioning on all4-parsed under strict ($n{=}262$, dropping $38$/$300$ items where some condition's strict parse fails) flips the sign to $+3.05$ pp. Both effects are real: under strict raw on $n{=}300$ items, CoT has $16$ fewer correct responses than direct on 32B ($-5.33$ pp = $169$ vs.\ $185$ correct); under strict all4-parsed on the smaller $n{=}262$ subset, the surviving CoT responses are slightly above direct accuracy. We do not present either as ``equivalence to direct''---the strict raw 90\% CI excludes zero on the harm side, and the strict all4 90\% CI upper bound of $+6.87$ pp is not within a $\pm 5$ pp sensitivity margin. \textbf{Second, on 7B the harm survives Holm correction under strict scoring.} Raw strict $-7.32$ pp, exact McNemar $p{=}0.003$, Holm-adjusted $p{=}0.012$ over the four-test primary multiplicity family---this is the only Holm-surviving test in the manuscript-declared primary family at $\alpha{=}0.05$. The same effect under permissive scoring is $-5.81$ pp ($p{=}0.019$, Holm $0.077$, does not survive). All four 7B CoT$-$direct point estimates are negative, all four 90\% CIs exclude zero, and the video-dependent subset analysis (\cref{app:vds}) is consistent with directional harm. \textbf{Third}, the answer-first control rules out a pure answer-position effect under permissive scoring (AF$-$direct $\approx 0$), but its scope narrows under strict on 7B (AF$-$direct $-3.79$ pp), so it does not isolate format-mediated effects from chain-generation burden; a non-CoT instruction matched on output length is future work.

\subsection{Is forced CoT video-conditioned?}
\label{sec:swap}

\begin{figure}[t]
\centering
\includegraphics[width=\linewidth]{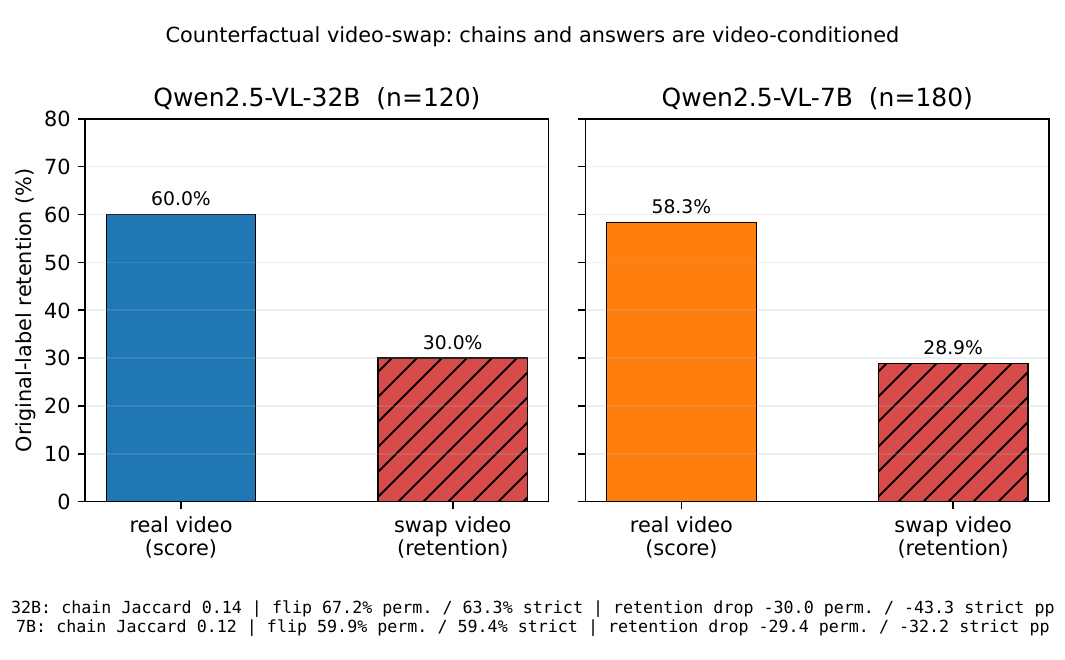}
\caption{Counterfactual video-swap probe. Bars show \emph{original-label retention under invalid video}: the fraction of probe items where the model's letter on the swapped video equals the gold letter for the \emph{original} video, computed over the full probe set ($n{=}120$/$180$) with non-parses counted as not-retained. The original gold label is not a valid score for the swap video; this metric is a refusal-vs-coincidental-match diagnostic. Jaccard and letter-flip are measured over paired CoT chains. \emph{Decoding is deterministic.} Under the boilerplate-chain prediction, Jaccard would be $\approx 1.0$ and letter-flip $\approx 0\%$; observed values are the opposite on this setup.}
\Description{Grouped bars for Qwen 32B and 7B compare real-video original-label retention against swapped-video retention under permissive and strict scoring, with annotations for chain Jaccard, letter-flip rate, and retention drops.}
\label{fig:swap}
\end{figure}

For each probe question we re-run the CoT condition with the real video replaced by a randomly chosen different video of the same task type (same duration bucket when available). We then measure: (a)~chain token-Jaccard between real and swap responses (content-word overlap, stopwords and common filler removed, \texttt{Answer:~X} lines stripped), (b)~final-letter flip rate over paired-parsed items, and (c)~\emph{original-label retention under invalid video}---a refusal-vs-coincidental-match diagnostic, not a valid benchmark score under the swapped video, decomposed into parsed-same / parsed-different / no-parse.

Results (\cref{fig:swap}): chain-Jaccard is $0.140$ (32B, $n{=}120$) / $0.119$ (7B, $n{=}180$). Under the permissive $(a)$--$(d)$ scorer the swap-parsed denominator is $117$/$178$ and the paired-parsed denominator is $116$/$177$; flip rate on the paired-parsed denominator is $67.2\%$/$59.9\%$ (recomputed end-to-end). Original-label retention on the full $n{=}120$/$180$ probe set drops from $60.0\% \rightarrow 30.0\%$ on 32B ($\Delta{=}{-}30.0$ pp) and $58.3\% \rightarrow 28.9\%$ on 7B ($\Delta{=}{-}29.4$ pp). The strict $(a)$--$(c)$ scorer collapses the swap-parsed denominator on 32B from $117$ to $57$---the dropped $60$ items are explicit refusals (e.g., ``the video does not match any option''), which the permissive scorer's fallback~$(d)$ coerces to a letter; we treat this collapse as evidence that strict scoring is the cleaner refusal/no-parse separator. Under strict, original-label retention drops by $43.3$ pp (32B) / $32.2$ pp (7B). The diagnostic value is in the \emph{drop} together with the Jaccard collapse, not in the absolute magnitude of post-swap retention (which is not an accuracy quantity since the original gold need not apply to the swapped video, so a uniform-random baseline is not meaningful here). \Cref{tab:strict-swap} lists every denominator and decomposes the swap pair into \emph{parsed-same / parsed-different / unparseable-under-row's-scorer}; the strict-row's $63/120$ unparseable count on 32B is the headline diagnostic for fallback~$(d)$ coercion. Under the boilerplate-chain prediction, Jaccard would be near $1.0$ and the letter-flip rate near $0\%$; we observe the opposite by an order of magnitude. The CoT chain on this setup is not input-invariant.

\paragraph{Chain-length check.} Low Jaccard is not a short-chain artifact. On the \emph{primary} (real-video) CoT condition, responses average $251$ whitespace tokens (median $239$, min $55$, max $413$) on 32B and $116$ tokens (median $96$, min $37$, max $438$) on 7B; the real-video response in the qualitative example below ($149$ tokens) sits inside this primary 32B range. The \emph{swap} response in the example ($49$ tokens) is shorter because it is an explicit refusal (``the video content does not match any option''), which is the typical strict-no-parse pattern \cref{tab:strict-swap}; the swap-population minimum can fall below the primary 32B minimum because refusals are short. The 32B real-video chains are on average $2.2\times$ longer than the 7B real-video chains, giving ample surface area for token overlap had the chains been boilerplate. The 7B chains being much shorter is consistent with the error-propagation reading we discuss in \cref{sec:discussion}.

\paragraph{Qualitative example.} \Cref{fig:example} shows a representative swap pair at $J{=}0.13$ on a \texttt{short}-duration \texttt{Action Recognition} question. The real video shows two Spider-Men characters; the 32B model reads their actions (pouring and drinking tea) and answers correctly (C, ``Drinking tea''). The swap video is a baseball match; the model explicitly refuses the MCQ because it observes that the video content does not match any answer option. The two chains share fewer than one content word in eight. A boilerplate-chain account would predict nearly identical chains: the model instead describes what it observes.

\begin{figure}[t]
\centering
\begin{minipage}{0.96\linewidth}\small
\textbf{Question (qid 097-1, Action Recognition, short):} \emph{What are the two Spider-Men doing in the video?}\\ (A) Eating cakes \ (B) Fighting monsters \ (C) Drinking tea \ (D) Sleeping

\vspace{2pt}
\textbf{Real video} (Spider-Men skit, ground-truth C)\\
\textit{``In the first few frames, we see two Spider-Men sitting at a table. One of them is holding a teapot and a cup\ldots\ The Spider-Man on the right is seen pouring tea into a cup\ldots\ Both Spider-Men are shown holding cups\ldots\ \textbf{Answer:~C. Drinking tea.}''}

\vspace{2pt}
\textbf{Swapped video} (same \texttt{Action Recognition} task type, different video: a baseball match)\\
\textit{``The provided video clips depict a baseball game and do not include any scenes of spider-men, eating cakes, fighting monsters, drinking tea, or sleeping. Therefore, none of the options A, B, C, or D are applicable to the content of the video. \textbf{Answer:~None of the above options apply.}''}

\vspace{2pt}
\textbf{Chain token-Jaccard $=0.13$.} The chain content tracks the video, not the prompt.
\end{minipage}
\caption{Representative CoT chain pair under the counterfactual-swap probe. Both responses are shortened for display; original lengths are $914$ and $283$ characters ($149$ and $49$ whitespace tokens). Decoding is greedy.}
\Description{Textual example showing the original Spider-Men question, a shortened real-video chain-of-thought answer ending in C, a shortened swapped baseball-video refusal, and chain token-Jaccard of 0.13.}
\label{fig:example}
\end{figure}

\subsection{How much visual information is used?}
\label{sec:deg}

\begin{figure}[t]
\centering
\includegraphics[width=\linewidth]{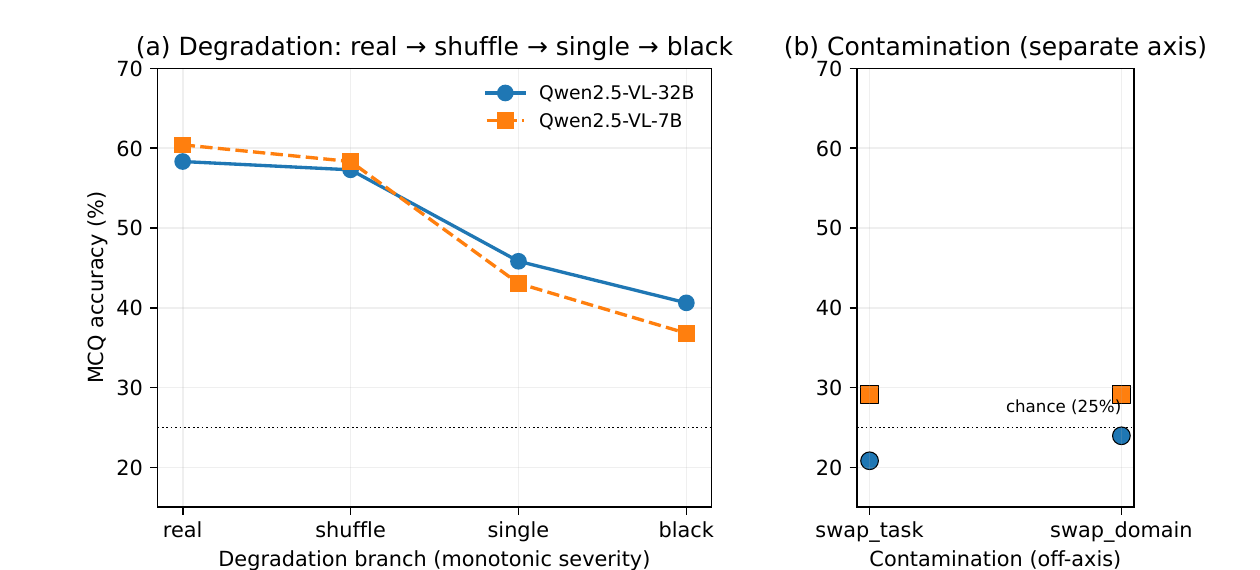}
\caption{Two-panel ladder. (a)~Four-rung \emph{degradation} branch on a single severity axis (\texttt{real} $\to$ \texttt{shuffle} $\to$ \texttt{single} $\to$ \texttt{black}); points are connected because the axis is monotonic. (b)~Two \emph{contamination} conditions (\texttt{swap\_task}, \texttt{swap\_domain}) on a separate axis; markers are unconnected because contamination is not on the same severity scale as degradation. Dotted line: $25\%$ four-way random. Both models use distinct markers and line styles for B/W readability.}
\Description{Two-panel line and marker plot. The degradation panel shows accuracy falling from real to shuffled to single-frame to black inputs for both models. The contamination panel shows lower accuracy for swap-task and swap-domain conditions, with a 25 percent random baseline.}
\label{fig:ladder}
\end{figure}

We evaluate the CoT condition under the four-rung degradation branch (\texttt{real} $\to$ \texttt{shuffle} $\to$ \texttt{single} $\to$ \texttt{black}, monotonically decreasing visual signal) plus two contamination conditions, \texttt{swap\_task} and \texttt{swap\_domain}. We treat the latter two as a different evaluation question (wrong video, not absent video) and do not assume they sit on the same scalar severity axis as the degradation branch.

\Cref{fig:ladder} and \Cref{tab:ladder} show the full per-rung numbers. Across the four-rung degradation branch alone, accuracy is strictly decreasing on both models (Spearman $\rho{=}{-}1.0$ on each; exact-permutation one-tailed $p{=}1/24{\approx}0.042$, $n{=}4$; \emph{Holm-adjusted $p{=}0.125$ over the four-test primary multiplicity family --- does not survive at $\alpha{=}0.05$, presented as a descriptive monotone pattern only}); chain token-Jaccard versus the real-chain co-decreases ($1.0 \to 0.36 \to 0.24 \to 0.24$ on 32B, $1.0 \to 0.32 \to 0.16 \to 0.15$ on 7B). Across all six conditions, ranks invert at the contamination tail: 32B has \texttt{swap\_task} $=$ $20.83\%$ $<$ \texttt{swap\_domain} $=$ $23.96\%$, and 7B ties both at $29.17\%$. Reporting Spearman across all six conditions therefore depends on assumed tie handling and on an ordering between \texttt{swap\_task} and \texttt{swap\_domain} that we do not endorse, so we omit a $p$-value at $n{=}6$. Three sub-patterns are visible. (i)~\texttt{shuffle} costs about $1$--$2$ pp of accuracy ($1.04$ on 32B, $2.09$ on 7B) but already drops chain-Jaccard to $\approx 0.32$ on 7B and $0.36$ on 32B, suggesting the CoT chain uses bag-of-frame information rather than strict temporal order, yet the surface text does change with frame order. (ii)~\texttt{single} costs about $12$--$18$ pp ($12.50$ on 32B; $17.36$ on 7B) and \texttt{black} costs about $18$--$24$ pp ($17.71$ on 32B; $23.61$ on 7B), so multi-frame dynamics carry real signal. (iii)~\texttt{swap\_task} and \texttt{swap\_domain} fall \emph{below} the primary \texttt{no\_video} floor on the same questions: re-evaluating \texttt{no\_video} on the ladder subset gives $38.5\%$ ($n{=}96$, 32B) / $40.3\%$ ($n{=}144$, 7B), so swap conditions are still $\sim15$ pp ($32$B) and $\sim11$ pp ($7$B) below same-subset \texttt{no\_video}. Chain-Jaccard versus the real-chain drops in lockstep with accuracy ($1.0 \to 0.13$ on 32B, $1.0 \to 0.11$ on 7B), mirroring rather than merely tracking the accuracy drop.

\begin{table}[t]
\centering\footnotesize
\caption{Visual-corruption ladder, both models. Four-rung \emph{degradation} branch (real$\to$shuffle$\to$single$\to$black) on a single severity axis; \emph{contamination} (swap-task, swap-domain) reported separately. $\rho_4$ = Spearman over the degradation branch only (rank-only test, scorer-invariant). Same-subset \texttt{no-video} accuracy on the $n{=}96/144$ ladder subset: $38.5\%$ (32B) / $40.3\%$ (7B).}
\label{tab:ladder}
\setlength{\tabcolsep}{4pt}
\begin{tabular}{@{}llcccc@{}}
\toprule
Model & cond & $n$ & acc \% & Jacc.\ vs.\ real & flip \% \\
\midrule
\multirow{4}{*}{\makecell[l]{32B \\ $\rho_4{=}{-}1.0$}}
& real          & $96$ & $58.33$ & $1.000$ & $0.0$ \\
& shuffle       & $96$ & $57.29$ & $0.360$ & $37.5$ \\
& single        & $96$ & $45.83$ & $0.236$ & $54.2$ \\
& black         & $96$ & $40.62$ & $0.235$ & $60.4$ \\
\cmidrule(lr){2-6}
& swap-task     & $96$ & $20.83$ & $0.144$ & $76.0$ \\
& swap-domain   & $96$ & $23.96$ & $0.130$ & $72.9$ \\
\midrule
\multirow{4}{*}{\makecell[l]{7B \\ $\rho_4{=}{-}1.0$}}
& real          & $144$ & $60.42$ & $1.000$ & $0.0$ \\
& shuffle       & $144$ & $58.33$ & $0.316$ & $26.4$ \\
& single        & $144$ & $43.06$ & $0.162$ & $52.8$ \\
& black         & $144$ & $36.81$ & $0.148$ & $61.8$ \\
\cmidrule(lr){2-6}
& swap-task     & $144$ & $29.17$ & $0.127$ & $65.3$ \\
& swap-domain   & $144$ & $29.17$ & $0.113$ & $61.1$ \\
\bottomrule
\end{tabular}
\end{table}

\section{Discussion}
\label{sec:discussion}

\paragraph{Grounding does not imply utility (in this instantiation).}
Chain-of-thought is often adopted under the implicit assumption that a plausible, input-conditioned rationale implies a more trustworthy final answer. On Video-MME with Qwen2.5-VL that assumption is not supported: on 32B, video-conditioned chains yield no statistically detected gain over direct answering under either scorer; on 7B, all four CoT$-$direct point estimates are negative, all 90\% CIs exclude zero, and raw-strict ($p{=}0.012$ Holm-adjusted) is the single test in the four-test primary family that survives Holm at $\alpha{=}0.05$. A plausible mechanism for the 7B harm is error propagation: the forced 5-step scaffold commits a smaller model to intermediate inferences whose errors survive into the final letter, while direct answering short-circuits the chain---we offer this only as a hypothesis. An auditing pipeline that checks whether a chain ``consults'' the visual input would have nothing to flag on our setup: the chains consult the input, yet still fail to help, and sometimes hurt.

\paragraph{What each probe catches.}
\Cref{tab:probe-ablation} sketches the conclusion that would be reached using each strict subset of our recipe. Without parse-rate-controlled scoring, the 32B raw $-2.33$ pp would be over-attributed to ``model harm'' instead of partly to format compliance. Without the swap probe, the boilerplate-chain prediction would not be testable at all. Without the four-rung ladder, the visual-signal monotonicity claim would rest only on aggregate Jaccard. The contribution is in their combination.

\begin{table}[t]
\centering\footnotesize
\caption{What each probe catches in this instantiation. Each row removes one probe and lists the single conclusion that becomes underdetermined (or wrong).}
\label{tab:probe-ablation}
\setlength{\tabcolsep}{4pt}
\begin{tabular}{@{}p{0.3\linewidth}p{0.6\linewidth}@{}}
\toprule
Drop & Conclusion that becomes underdetermined / wrong \\
\midrule
parsed-only conditioning & The 32B raw-strict $-5.33$ pp harm versus all4-strict $+3.05$ pp would collapse into a single uninterpretable summary, and the parse-rate selection mechanism would be invisible. \\
swap probe & The boilerplate-chain prediction (Jaccard $\approx 1.0$, flip $\approx 0\%$) would not be testable on this setup at all. \\
four-rung ladder & Graded grounding (accuracy and chain-Jaccard both decrease monotonically over the degradation branch) would rest on aggregate Jaccard alone. \\
strict-vs-permissive scorer & The 32B parsed-only sign \emph{flip} between scorers ($-2.01$ permissive vs.\ $+3.05$ strict on $n{=}262$) and the change in 7B Holm-survival ($p{=}0.077$ permissive vs.\ $p{=}0.012$ strict) would both be hidden under a single-scorer report. \\
\bottomrule
\end{tabular}
\end{table}

\paragraph{Lessons for evaluation design.}
We frame the contribution as a \emph{recipe} (a worked instantiation future work can port) rather than a settled \emph{protocol}. For evaluators of multimodal generation: (i)~report parse-rate alongside accuracy and report \emph{both} a permissive and a strict scorer so any sign flip is visible; (ii)~report exact paired tests with bootstrap CIs and Holm-correct over a pre-declared primary multiplicity family; (iii)~separate input-conditioning claims from accuracy claims with at least one input-substitution probe; (iv)~when reporting graded grounding, an ordered degradation \emph{branch} produces a falsifiable Spearman test, but small numbers of rungs ($n{=}4$ here) admit only modest one-tailed permutation $p$-values that do not survive Holm in a four-test family, and contamination conditions belong on a separate axis.

\section{Limitations}
\label{sec:limitations}

\begin{enumerate}\setlength\itemsep{0em}
\item \textbf{Scope and external validity.} Results cover $n{=}300$ / $396$ of 2700 Video-MME questions on Qwen2.5-VL at two scales (32B in \texttt{bitsandbytes} NF4 4-bit, 7B in \texttt{bf16}) under a single forced-rationale prompt; full-benchmark evaluation, alternate quantization (\texttt{bf16} / AWQ), a third scale, cross-family confirmation (Llava-Video, InternVL-Video, Gemini), and prompt-style sweeps are deferred. The error-propagation reading in \cref{sec:discussion} is offered as a conjecture; no manual error taxonomy was performed.
\item \textbf{Custom letter extractor not externally validated.} Our regex cascade is not the official Video-MME evaluator; we mitigate by reporting both permissive $(a)$--$(d)$ and strict $(a)$--$(c)$ scorers on every primary table, but item-level agreement with the official Video-MME evaluator is deferred and is the most important single follow-up. Until that check is run, no claim in this paper should be read as scorer-independent.
\end{enumerate}

\section{Conclusion}
\label{sec:conclusion}

We asked a simple question: does forced chain-of-thought help a video-language model answer multiple-choice questions? On Qwen2.5-VL with Video-MME, the answer is \emph{no}. The chains the model writes do depend on the video---swap the input video and the chain content collapses---but that input-conditioning does not translate into better final answers. On the larger 32B model, we detect no difference between CoT and direct prompting (we explicitly do not claim equivalence; the strict raw 90\% CI excludes zero on the harm side, and the strict all-four-parsed CI is wide). On the smaller 7B model, CoT actually hurts; the drop survives multiplicity correction under our primary scoring choice.

We caveat this in the obvious places. The result is one model family on subsets of one benchmark, with a custom regex scorer that has not been cross-checked against the official Video-MME evaluator. We chose the strict scorer as primary after seeing the data, and the 7B significance result depends on that choice; we report both scorers everywhere so the reader can see the dependence. The specific narrative ``video-conditioned chains improve accuracy'' fails here, but we make no claim about other benchmarks, scales, or prompt styles.

What we contribute is a small reusable evaluation recipe: report parse rate alongside accuracy under both a strict and a permissive scorer; add an input-substitution probe to test whether chains are truly input-conditioned; and use an ordered visual-degradation branch to ask how much visual signal the chain actually uses. Each probe targets a question the standard ``CoT helps / doesn't help'' framing tends to leave unanswered. The raw responses, item lists, prompt hashes, and a single recomputation script (\cref{app:artifact}) will be released with the supplementary material so every number in the paper can be reproduced without any new model inference; the most useful follow-ups are item-level agreement with the official Video-MME evaluator, validation on the remaining $\sim 2{,}300$ Video-MME items, a second model family, and a manual error taxonomy of the 7B CoT failures.

\bibliographystyle{ACM-Reference-Format}
\bibliography{references}

@inproceedings{weicot2022,
  author    = {Jason Wei and Xuezhi Wang and Dale Schuurmans and Maarten Bosma and
               Brian Ichter and Fei Xia and Ed H. Chi and Quoc V. Le and Denny Zhou},
  title     = {Chain-of-Thought Prompting Elicits Reasoning in Large Language Models},
  booktitle = {Advances in Neural Information Processing Systems 35 (NeurIPS 2022)},
  year      = {2022},
  url       = {https://arxiv.org/abs/2201.11903}
}

@inproceedings{kojimazsot2022,
  author    = {Takeshi Kojima and Shixiang Shane Gu and Machel Reid and
               Yutaka Matsuo and Yusuke Iwasawa},
  title     = {Large Language Models are Zero-Shot Reasoners},
  booktitle = {Advances in Neural Information Processing Systems 35 (NeurIPS 2022)},
  year      = {2022},
  url       = {https://arxiv.org/abs/2205.11916}
}

@inproceedings{turpin2023,
  author    = {Miles Turpin and Julian Michael and Ethan Perez and
               Samuel R. Bowman},
  title     = {Language Models Don't Always Say What They Think:
               Unfaithful Explanations in Chain-of-Thought Prompting},
  booktitle = {Advances in Neural Information Processing Systems 36 (NeurIPS 2023)},
  year      = {2023},
  url       = {https://arxiv.org/abs/2305.04388}
}

@article{lanham2023,
  author    = {Tamera Lanham and Anna Chen and Ansh Radhakrishnan and Benoit Steiner and Carson Denison and Danny Hernandez and Dustin Li and Esin Durmus and Evan Hubinger and Jackson Kernion and Kamile Lukosiute and Karina Nguyen and Newton Cheng and Nicholas Joseph and Nicholas Schiefer and Oliver Rausch and Robin Larson and Sam McCandlish and Sandipan Kundu and Saurav Kadavath and Shannon Yang and Thomas Henighan and Timothy Maxwell and Timothy Telleen-Lawton and Tristan Hume and Zac Hatfield-Dodds and Jared Kaplan and Jan Brauner and Samuel R.~Bowman and Ethan Perez},
  title     = {Measuring Faithfulness in Chain-of-Thought Reasoning},
  journal   = {arXiv preprint arXiv:2307.13702},
  year      = {2023},
  url       = {https://arxiv.org/abs/2307.13702}
}

@article{madaan2022text,
  author    = {Aman Madaan and Amir Yazdanbakhsh},
  title     = {Text and Patterns: For Effective Chain of Thought, It Takes Two to Tango},
  journal   = {arXiv preprint arXiv:2209.07686},
  year      = {2022},
  url       = {https://arxiv.org/abs/2209.07686}
}

@inproceedings{videomme2024,
  author    = {Chaoyou Fu and Yuhan Dai and Yongdong Luo and Lei Li and Shuhuai Ren and Renrui Zhang and Zihan Wang and Chenyu Zhou and Yunhang Shen and Mengdan Zhang and Peixian Chen and Yanwei Li and Shaohui Lin and Sirui Zhao and Ke Li and Tong Xu and Xiawu Zheng and Enhong Chen and Caifeng Shan and Ran He and Xing Sun},
  title     = {{Video-MME}: The First-Ever Comprehensive Evaluation Benchmark of Multi-modal {LLMs} in Video Analysis},
  booktitle = {Proceedings of the IEEE/CVF Conference on Computer Vision and Pattern Recognition (CVPR)},
  pages     = {24108--24118},
  year      = {2025},
  doi       = {10.1109/CVPR52734.2025.02245},
  url       = {https://arxiv.org/abs/2405.21075}
}

@article{qwen25vl2025,
  author    = {Shuai Bai and Keqin Chen and Xuejing Liu and Jialin Wang and Wenbin Ge and Sibo Song and Kai Dang and Peng Wang and Shijie Wang and Jun Tang and Humen Zhong and Yuanzhi Zhu and Mingkun Yang and Zhaohai Li and Jianqiang Wan and Pengfei Wang and Wei Ding and Zheren Fu and Yiheng Xu and Jiabo Ye and Xi Zhang and Tianbao Xie and Zesen Cheng and Hang Zhang and Zhibo Yang and Haiyang Xu and Junyang Lin},
  title     = {{Qwen2.5-VL} Technical Report},
  journal   = {arXiv preprint arXiv:2502.13923},
  year      = {2025},
  url       = {https://arxiv.org/abs/2502.13923}
}

@misc{qwen25vl32b2025,
  author    = {{Qwen Team}},
  title     = {{Qwen2.5-VL-32B}: Smarter and Lighter},
  year      = {2025},
  howpublished = {Qwen blog, March 24, 2025},
  url       = {https://qwenlm.github.io/blog/qwen2.5-vl-32b/}
}

@article{schuirmann1987tost,
  author    = {Donald J. Schuirmann},
  title     = {A Comparison of the Two One-Sided Tests Procedure and the Power Approach for Assessing the Equivalence of Average Bioavailability},
  journal   = {Journal of Pharmacokinetics and Biopharmaceutics},
  volume    = {15},
  number    = {6},
  pages     = {657--680},
  year      = {1987},
  doi       = {10.1007/BF01068419}
}

@article{lakens2017tost,
  author    = {Dani{\"e}l Lakens},
  title     = {Equivalence Tests: A Practical Primer for $t$ Tests, Correlations, and Meta-Analyses},
  journal   = {Social Psychological and Personality Science},
  volume    = {8},
  number    = {4},
  pages     = {355--362},
  year      = {2017},
  doi       = {10.1177/1948550617697177}
}

@article{zhang2023multimodalcot,
  author    = {Zhuosheng Zhang and Aston Zhang and Mu Li and Hai Zhao and George Karypis and Alex Smola},
  title     = {Multimodal Chain-of-Thought Reasoning in Language Models},
  journal   = {Transactions on Machine Learning Research},
  year      = {2024},
  url       = {https://arxiv.org/abs/2302.00923}
}

@article{mcnemar1947note,
  author    = {Quinn McNemar},
  title     = {Note on the Sampling Error of the Difference Between Correlated Proportions or Percentages},
  journal   = {Psychometrika},
  volume    = {12},
  number    = {2},
  pages     = {153--157},
  year      = {1947},
  doi       = {10.1007/BF02295996}
}

@article{fagerland2013recommended,
  author    = {Morten W. Fagerland and Stian Lydersen and Petter Laake},
  title     = {The {McNemar} Test for Binary Matched-Pairs Data: Mid-$p$ and Asymptotic Are Better Than Exact Conditional},
  journal   = {BMC Medical Research Methodology},
  volume    = {13},
  pages     = {91},
  year      = {2013},
  doi       = {10.1186/1471-2288-13-91}
}

@book{efron1993bootstrap,
  author    = {Bradley Efron and Robert J. Tibshirani},
  title     = {An Introduction to the Bootstrap},
  publisher = {Chapman \& Hall/CRC},
  series    = {Monographs on Statistics and Applied Probability},
  number    = {57},
  year      = {1993}
}

@article{holm1979simple,
  author    = {Sture Holm},
  title     = {A Simple Sequentially Rejective Multiple Test Procedure},
  journal   = {Scandinavian Journal of Statistics},
  volume    = {6},
  number    = {2},
  pages     = {65--70},
  year      = {1979}
}

@inproceedings{li2024image,
  author    = {Lei Li},
  title     = {{CPSeg}: Finer-Grained Image Semantic Segmentation via Chain-of-Thought Language Prompting},
  booktitle = {Proceedings of the IEEE/CVF Winter Conference on Applications of Computer Vision (WACV)},
  pages     = {513--522},
  year      = {2024},
  url       = {https://openaccess.thecvf.com/content/WACV2024/html/Li_CPSeg_Finer-Grained_Image_Semantic_Segmentation_via_Chain-of-Thought_Language_Prompting_WACV_2024_paper.html}
}

@inproceedings{li2025human,
  author    = {Lei Li and Sen Jia and Jianhao Wang and Zhongyu Jiang and Feng Zhou and Ju Dai and Tianfang Zhang and Zongkai Wu and Jenq-Neng Hwang},
  title     = {Human Motion Instruction Tuning},
  booktitle = {Proceedings of the IEEE/CVF Conference on Computer Vision and Pattern Recognition (CVPR)},
  year      = {2025},
  url       = {https://arxiv.org/abs/2411.16805}
}

@inproceedings{yao2025countllm,
  author    = {Ziyu Yao and Xuxin Cheng and Zhiqi Huang and Lei Li},
  title     = {{CountLLM}: Towards Generalizable Repetitive Action Counting via Large Language Model},
  booktitle = {Proceedings of the IEEE/CVF Conference on Computer Vision and Pattern Recognition (CVPR)},
  year      = {2025},
  url       = {https://arxiv.org/abs/2503.17690}
}

@inproceedings{cai2025role,
  author    = {Chengkun Cai and Xu Zhao and Haoliang Liu and Zhongyu Jiang and Tianfang Zhang and Zongkai Wu and Jenq-Neng Hwang and Lei Li},
  title     = {The Role of Deductive and Inductive Reasoning in Large Language Models},
  booktitle = {Proceedings of the 63rd Annual Meeting of the Association for Computational Linguistics (Volume 1: Long Papers)},
  pages     = {16780--16790},
  year      = {2025},
  publisher = {Association for Computational Linguistics},
  doi       = {10.18653/v1/2025.acl-long.820},
  url       = {https://aclanthology.org/2025.acl-long.820/}
}

@inproceedings{lan-etal-2025-attention,
  author    = {Tian Lan and Jinyuan Xu and Xue He and Jenq-Neng Hwang and Lei Li},
  title     = {Attention Consistency for {LLM}s Explanation},
  booktitle = {Findings of the Association for Computational Linguistics: EMNLP 2025},
  pages     = {1736--1750},
  year      = {2025},
  address   = {Suzhou, China},
  publisher = {Association for Computational Linguistics},
  doi       = {10.18653/v1/2025.findings-emnlp.91},
  url       = {https://aclanthology.org/2025.findings-emnlp.91/}
}

@article{ji2025mrag,
  author    = {Yuelyu Ji and Wuwei Lan and Patrick Ng},
  title     = {{MRAG-Suite}: A Diagnostic Evaluation Platform for Visual Retrieval-Augmented Generation},
  journal   = {arXiv preprint arXiv:2509.24253},
  year      = {2025},
  url       = {https://arxiv.org/abs/2509.24253}
}

@article{luo2024bigbench,
  author    = {Hanjun Luo and Haoyu Huang and Ziye Deng and Xinfeng Li and Hewei Wang and Yingbin Jin and Yang Liu and Wenyuan Xu and Zuozhu Liu},
  title     = {{BIGbench}: A Unified Benchmark for Evaluating Multi-Dimensional Social Biases in Text-to-Image Models},
  journal   = {arXiv preprint arXiv:2407.15240},
  year      = {2024},
  url       = {https://arxiv.org/abs/2407.15240}
}

@article{luo2026biasig,
  author    = {Hanjun Luo and Zhimu Huang and Haoyu Huang and Ziye Deng and Ruizhe Chen and Xinfeng Li and Zuozhu Liu and Hanan Salam},
  title     = {{BiasIG}: Benchmarking Multi-Dimensional Social Biases in Text-to-Image Models},
  journal   = {arXiv preprint arXiv:2604.11934},
  year      = {2026},
  note      = {Accepted by IJCNN 2026},
  url       = {https://arxiv.org/abs/2604.11934}
}

@article{luo2024faintbench,
  author    = {Hanjun Luo and Ziye Deng and Ruizhe Chen and Zuozhu Liu},
  title     = {{FAIntbench}: A Holistic and Precise Benchmark for Bias Evaluation in Text-to-Image Models},
  journal   = {arXiv preprint arXiv:2405.17814},
  year      = {2024},
  note      = {Accepted by ICML DMLR 2024},
  url       = {https://arxiv.org/abs/2405.17814}
}

@article{zhuang2026preregistering,
  author    = {Zexin Zhuang and Yanhang Li and Zhichao Fan},
  title     = {Pre-Registering the Detectable Effect: A Paired-{MDE} Budget for 4-Bit Quantization Benchmarks, with a Pilot Audit},
  journal   = {arXiv preprint arXiv:2605.28873},
  year      = {2026},
  url       = {https://arxiv.org/abs/2605.28873}
}

@article{li2026safetyrepro,
  author    = {Yanhang Li and Zhichao Fan and Zexin Zhuang},
  title     = {{SafetyRepro}: Configuration-Conditional Rank Instability on Alignment Benchmarks},
  journal   = {arXiv preprint arXiv:2605.25492},
  year      = {2026},
  url       = {https://arxiv.org/abs/2605.25492}
}

@article{li2026auditing,
  author    = {Yanhang Li and Zhichao Fan and Zexin Zhuang},
  title     = {Auditing Reasoning-Trace Memorization Claims after Unlearning with Head-Conditioned Canaries},
  journal   = {arXiv preprint arXiv:2605.18891},
  year      = {2026},
  url       = {https://arxiv.org/abs/2605.18891}
}

@misc{chen2026doesragknowretrieval,
  author        = {Yihang Chen and Pin Qian and Su Wang and Sipeng Zhang and Huan Xu and Shuhuai Lin and Xinpeng Wei},
  title         = {Does {RAG} Know When Retrieval Is Wrong? Diagnosing Context Compliance under Knowledge Conflict},
  year          = {2026},
  eprint        = {2605.14473},
  archivePrefix = {arXiv},
  primaryClass  = {cs.CL},
  url           = {https://arxiv.org/abs/2605.14473}
}

@misc{qian2026relevantwarrantedevidenceforcecalibration,
  author        = {Pin Qian and Su Wang and Xiaoyuan Wang and Yihang Chen and Wenxuan Xu and Qiaolin Yu and Shuhuai Lin and Sipeng Zhang and Junxian You and Xinpeng Wei},
  title         = {Relevant Is Not Warranted: Evidence-Force Calibration for Cited {RAG}},
  year          = {2026},
  eprint        = {2605.28044},
  archivePrefix = {arXiv},
  primaryClass  = {cs.AI},
  url           = {https://arxiv.org/abs/2605.28044}
}

\appendix
\section{Subtitle ablation (32B only)}
\label{app:subtitle}

When video subtitles (SRT) are available, we can inject them as a prompt prefix. Of the 300 primary 32B questions, only $n{=}61$ have matching SRT files; the ablation is therefore underpowered and is reported only as a small descriptive supplement. On those paired questions, adding subtitles moves direct raw accuracy by $+4.92$ pp ($70.5\%{\to}75.4\%$) and CoT raw by $+4.92$ pp ($63.93\%{\to}68.85\%$); the all-four-parsed CoT delta on the matched parsed subset is $+2.77$ pp. We do not attempt to compute CIs on $n{=}61$ paired and we make no claim about ``primary information channel''---the comparison mixes a subtitle-available subset with the primary subset and is exploratory.

\section{Parse-rate observation}
\label{app:parse}

Parse rates by condition under both scorers (fraction of responses for which the letter extractor returned a match), recomputed by \texttt{recompute\_canonical.py}:
\begin{center}\small
\begin{tabular}{@{}lcccc@{}}
\toprule
Model & direct & CoT & answer-first & no-video \\
\midrule
32B (permissive $a$--$d$) & $100.00\%$ & $99.67\%$ & $100.00\%$ & $100.00\%$ \\
32B (strict $a$--$c$)   & $100.00\%$ & $87.33\%$ & $100.00\%$ & $100.00\%$ \\
7B  (permissive $a$--$d$) & $100.00\%$ & $99.49\%$ & $99.24\%$  & $100.00\%$ \\
7B  (strict $a$--$c$)   & $100.00\%$ & $95.96\%$ & $90.66\%$  & $100.00\%$ \\
\bottomrule
\end{tabular}
\end{center}
The strict-vs-permissive gap on 32B CoT ($87.3\%$ vs.\ $99.7\%$) is the parser-sensitivity number for the \emph{primary direct-vs-CoT condition} and the all-four-parsed estimand: rule~$(d)$ in the permissive cascade rescues many CoT responses that finish with a free-form letter mention but no \texttt{Answer:~X} tag. The corresponding headline number for the \emph{swap probe} is the swap-parsed collapse from $117/120$ permissive to $57/120$ strict on 32B (NP-swap = $63/120$, see \cref{tab:strict-swap}); the two sensitivity numbers are not the same quantity.

\section{Per-task-type CoT gain/hurt}
\label{app:bytask}

\begin{figure}[h]
\centering
\includegraphics[width=\linewidth]{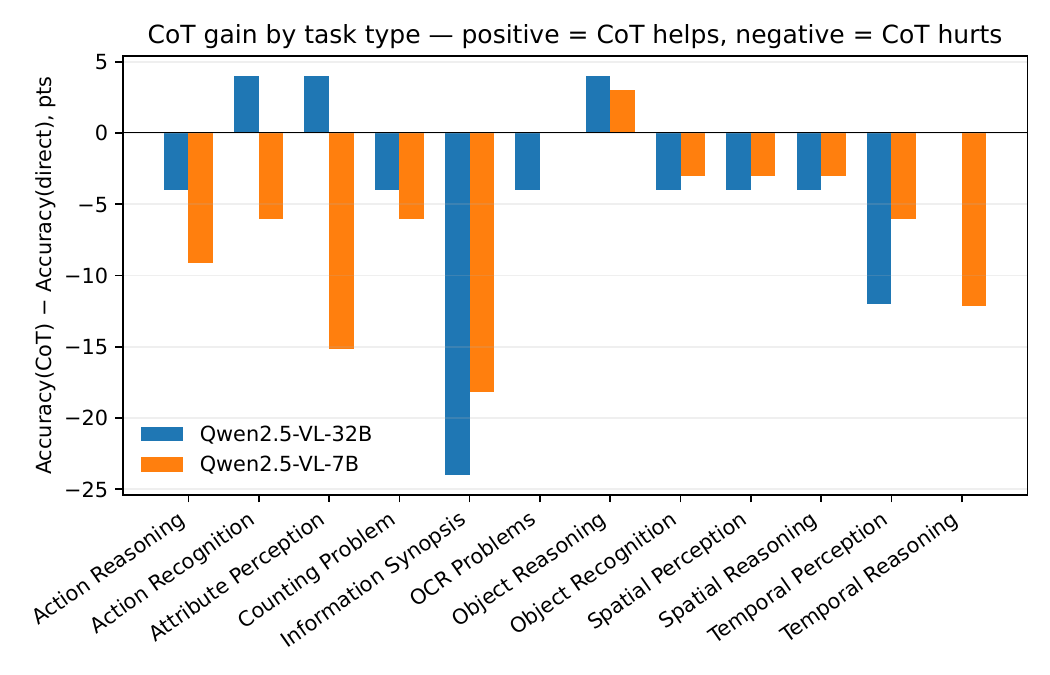}
\caption{CoT$-$direct accuracy delta by Video-MME task type. Bars below zero indicate CoT hurts; above zero, CoT helps.}
\Description{Bar chart of per-task CoT-minus-direct accuracy deltas by Video-MME task type, with bars above and below zero for the two model sizes.}
\label{fig:bytask}
\end{figure}

\Cref{fig:bytask} decomposes CoT$-$direct by Video-MME task type. Patterns are noisy at $n{=}25$--$33$ per bucket but broadly consistent with the paired results.

\section{Scorer robustness on the primary direct-vs-CoT comparison}
\label{app:strict-primary}

\Cref{tab:primary} already reports both scorers' all-four-parsed rows. \Cref{tab:7b-pair} reports the 7B \emph{pair-parsed} estimand (direct$\cap$CoT---items where only direct and CoT need to parse, regardless of AF and no-video) under the permissive scorer, and the all-four-parsed estimand under both scorers. We omit the strict pair-parsed (direct$\cap$CoT) row: under our canonical strict (a)--(c) extractor it would require item-level recomputation we defer to the artifact bundle. The 7B CoT$-$direct point estimate is in the range $[{-}5.88, {-}5.54]$ pp across the three reported 7B cells (perm-all4, perm-d$\cap$C, strict-all4), and all three 90\% CIs are entirely below zero. The 32B parsed-only sign instability between scorers (permissive $-2.01$; strict $+3.05$) is the substantive scorer-sensitivity finding and is already in Table~\ref{tab:primary}; we do not duplicate the 32B scorer comparison here.

\begin{table}[h]
\centering\footnotesize
\setlength{\tabcolsep}{3pt}
\caption{7B direct-vs-CoT under pair-parsed (direct$\cap$CoT) and all-four-parsed estimands. Permissive (a)--(d) is reported on both estimands; strict (a)--(c) on the all-four-parsed estimand only (see prose).}
\label{tab:7b-pair}
\begin{tabular}{@{}llcccc@{}}
\toprule
Scorer & Subset & $n$ & dir / CoT & $\Delta$ & 90\% CI ($p$) \\
\midrule
perm.\ ($a$-$d$) & all4   & $391$ & $59.34/53.45$ & $-5.88$ & $[-9.72, -2.05]$ ($0.018$) \\
perm.\ ($a$-$d$) & d$\cap$C & $394$ & $59.14/53.55$ & $-5.58$ & $[-9.39, -1.78]$ ($0.025$) \\
strict ($a$-$c$) & all4   & $343$ & $59.77/54.23$ & $-5.54$ & $[-9.62, -1.46]$ ($0.034$) \\
\bottomrule
\end{tabular}
\end{table}

\section{Strict-letter robustness on the swap probe}
\label{app:strict-swap}

We re-score the counterfactual swap (Probe 2) under the strict letter extractor and decompose each pair into parsed-same\,/\,parsed-different\,/\,strict-no-parse. The number of swap responses parseable by the permissive scorer but not by strict is $117-57=60$ on 32B and $178-162=16$ on 7B; the strict no-parse-on-swap count (the complement of strict-parsed) is $63/120$ on 32B and $18/180$ on 7B. These items are refusals (e.g.,~``the video does not match any option'') that fallback~$(d)$ coerces to a letter. The drop in original-label retention is preserved on both models and is sharper under strict (\cref{tab:strict-swap}). All numbers are recomputed by \texttt{recompute\_canonical.py} from \texttt{counterfactual.jsonl}.

\begin{table}[h]
\centering\footnotesize
\setlength{\tabcolsep}{4pt}
\caption{Swap-probe metrics under permissive vs.\ strict scoring. \emph{swap-prs}: swap responses parseable under the row's scorer (out of $n{=}120$/$180$); the complement (e.g., $63/120$ strict-32B) is explicit refusals that fallback~$(d)$ coerces into letters. \emph{pair-flip}: paired-parsed items whose letter changed under swap. \emph{retain}: original-label retention on real$\to$swap (full $n$, no-parse counted as not-retained).}
\label{tab:strict-swap}
\begin{tabular}{@{}llcccc@{}}
\toprule
Model & Scorer & swap-prs & pair-flip & retain (real$\to$swap) & $\Delta$ \\
\midrule
\multirow{2}{*}{32B}
 & perm.\ ($a$-$d$) & $117/120$ & $78/116$ & $60.0\to 30.0$ & $-30.0$ \\
 & strict ($a$-$c$) & $57/120$ & $31/49$ & $58.3\to 15.0$ & $-43.3$ \\
\midrule
\multirow{2}{*}{7B}
 & perm.\ ($a$-$d$) & $178/180$ & $106/177$ & $58.3\to 28.9$ & $-29.4$ \\
 & strict ($a$-$c$) & $162/180$ & $95/160$ & $57.8\to 25.6$ & $-32.2$ \\
\bottomrule
\end{tabular}
\end{table}

\section{Full prompt templates}
\label{app:prompts}

Each condition uses a single, fixed prompt template; only the trailing instruction differs across conditions. Below we show the user-side message for each (the system prompt is empty across conditions to avoid format drift):

\noindent\textbf{direct.} \emph{``Watch the video and answer the multiple-choice question. Question: \{question\}. Options: A) \{a\}, B) \{b\}, C) \{c\}, D) \{d\}. Answer with only the letter (A, B, C, or D).''}

\noindent\textbf{CoT.} \emph{``Watch the video and answer the multiple-choice question. Question: \{question\}. Options: A) \{a\}, B) \{b\}, C) \{c\}, D) \{d\}. Reason step by step in at least 5 numbered steps, citing a frame for each step. Conclude with `Answer: X' on its own line.''}

\noindent\textbf{answer-first.} \emph{``Watch the video and answer the multiple-choice question. Question: \{question\}. Options: A) \{a\}, B) \{b\}, C) \{c\}, D) \{d\}. First write `Answer: X' on its own line. Then justify the answer in at least 5 numbered steps, each citing a frame.''}

\noindent\textbf{no-video.} Same as \emph{direct} but with no video frames passed to the model; only the question and options are visible.

For \emph{swap\_task} and \emph{swap\_domain} (Probes 2 and 3), the prompt is identical to the CoT condition; only the video input changes. The full prompt JSON, including system-prompt text and stop-token configuration, will be in the supplementary recomputation script at \texttt{scripts/prompts.py}.

\noindent\textbf{Domain.} ``Domain'' refers to Video-MME's coarse-category partition (Knowledge, Film \& Television, Sports, Life Record, Multilingual, etc.), as listed in the benchmark metadata. \texttt{swap\_domain} draws a different video from a \emph{different} top-level domain than the original; \texttt{swap\_task} draws from the same task type (e.g., \emph{Action Recognition}) regardless of domain.

\section{Video-dependent subset analysis}
\label{app:vds}

We restrict the primary paired comparison to questions where the \texttt{no\_video} condition got the wrong answer---i.e.\ the question is not solvable from question/options alone (though we cannot claim it is fully video-essential).

\begin{center}\footnotesize
\setlength{\tabcolsep}{4pt}
\begin{tabular}{@{}lcccc@{}}
\toprule
Model & paired $n$ & CoT$-$direct & AF$-$direct & no-video acc \\
\midrule
32B & $174$ & $+0.57$ & $+4.02$ & $0\%$ (by const.) \\
7B  & $250$ & $-2.40$ & $0.00$  & $0\%$ \\
\bottomrule
\end{tabular}
\end{center}
The CoT non-gain at 32B and harm at 7B both persist on the video-dependent subset; the 7B point estimate is smaller in magnitude on this subset ($-2.40$ pp on $n{=}250$) than on the full paired sample ($-7.32$ pp raw-strict / $-5.81$ pp raw-permissive / $-5.54$ pp all4-strict from \cref{tab:primary}), consistent with some of the full-set 7B harm arising from easier items where CoT introduces error into a question the model could otherwise solve.

\section{Artifact bundle}
\label{app:artifact}

The camera-ready supplementary artifact bundle will contain:
\begin{itemize}\setlength\itemsep{0em}
\item \textbf{Raw responses (JSONL):} eight files per model (direct, CoT, AF, no-video, counterfactual, degradation, direct-subtitled, CoT-subtitled), one record per question, with fields \emph{question\_id}, \emph{task\_type}, \emph{duration\_bucket}, \emph{gold}, \emph{response}, \emph{prompt\_hash}, plus a SHA-256 manifest.
\item \textbf{One recomputation script} (Python 3.11+, no GPU, $<\!30$~s) that reproduces every number in \cref{tab:primary,tab:7b-pair,tab:strict-swap,tab:ladder}, the per-task table (\cref{app:bytask}), the parse-rate table (\cref{app:parse}), and the swap-probe extras file from the raw JSONL.
\item \textbf{Item lists:} the $300$/$396$ primary, $120$/$180$ swap, $96$/$144$ ladder, and $61$ subtitle question IDs, plus the swap-pair mapping (\emph{seed} $=0$).
\item \textbf{Prompt templates} (verbatim, see also \cref{app:prompts}) and the deterministic \emph{plot script} that reproduces \cref{fig:swap,fig:ladder} and the by-task figure.
\item \textbf{Recomputed canonical numbers} (the script's expected output) so reviewers can diff against their local rerun.
\end{itemize}
Once released, a reader will be able to verify every table and figure number from the JSONL using one Python invocation, \emph{without any VLM forward passes}; regenerating the JSONL itself requires Qwen2.5-VL-7B/32B inference, the Video-MME videos, and the \emph{seed} $=0$ configurations of \cref{sec:setup,app:prompts}.

\end{document}